# Model Checking in Medical Imaging for Tumor Detection and Segmentation

1st Elhoucine Elfatimi*
Department of Pathology Laboratory Medicine
University of California
Irvine, CA, USA
eelfatim@uci.edu

2nd Lahcen El Fatimi
Department of Computer and Software Engineering
Polytechnique Montreal
University of Montreal, Montreal, QC, Canada
lahcen.elfatimi@polymtl.ca

*Abstract*—Recent advancements in model checking have demonstrated significant potential across diverse applications, particularly in signal and image analysis. Medical imaging stands out as a critical domain where model checking can be effectively applied to design and evaluate robust frameworks. These frameworks facilitate automatic and semi-automatic delineation of regions of interest within images, aiding in accurate segmentation. This paper provides a comprehensive analysis of recent works leveraging spatial logic to develop operators and tools for identifying regions of interest, including tumorous and non-tumorous areas. Additionally, we examine the challenges inherent to spatial model-checking techniques, such as variability in ground truth data and the need for streamlined procedures suitable for routine clinical practice.

*Index Terms*—Model Checking, Segmentation, medical images, tumor.

## I. INTRODUCTION

Model checking is the process of verifying whether a given structure satisfies a specified logical formula. This concept is general and applies to a wide range of logics and system designs. A fundamental model-checking problem involves determining whether a propositional logic equation is satisfied by a given structure. Model checking is most commonly applied to hardware designs. For software systems, due to undecidability, the methodology cannot be fully algorithmic and may fail to either prove or disprove a given property.

Model checking plays a critical role across various applications, serving purposes such as ensuring the correctness of system properties and minimizing errors in software under development. Traditional model checking typically consists of three major steps:

- Formal Model of the System: This step involves creating a formal representation of the system in a language compatible with the model checker's requirements.
- Specification of System Properties: A specific property of the system is defined for verification. This translates into a question about the system's behavior that the model checker is expected to answer.
- Verification by the Model Checker: The model checker evaluates whether the specified property is satisfied. If the property cannot be verified, a counterexample is generated to identify the source of the error in the simulation model.

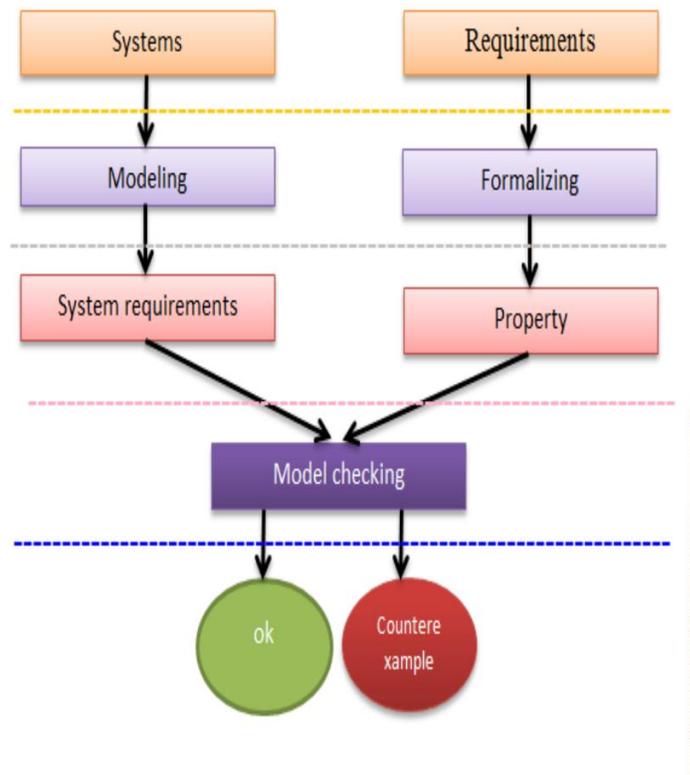

Figure 1: A typical model-checking workflow

Model checking has numerous applications, particularly in medical image analysis, such as image segmentation, contouring, filtering, and classification. These computations are closely tied to the spatial features of images. For example, computer-aided diagnosis focuses on classifying specific areas in an image that may correlate with a disease [1]. Similarly, image segmentation aims to identify regions within an image that exhibit specific features [2], such as tumors or lesions, depending on the application. Another approach involves forming contours around target areas corresponding to organs at risk, commonly used in radiotherapy [3]. Indicator-based schemes are also employed to compute metrics from images, aiding diagnostic procedures and enhancing understanding of disease-specific characteristics. These indicators can also be extended to monitor treatments and predict outcomes [4], with mean diffusivity serving as one such example [5].

This review paper aims to comprehensively understand and assess prior research on model checking in medical imaging, with a primary focus on segmentation-based approaches. Medical imaging encompasses multiple modalities, with magnetic resonance (MR) imaging being qualitatively assessed in clinical and routine settings. Typically, image contrast serves as a marker for the presence of hyperintense tissue [6]. Manual segmentation of regions of interest is labor-intensive, time-consuming, error-prone, and subjective. To address these limitations, automatic and semi-automatic segmentation algorithms have been developed. These methods are more accurate and reliable for isolating tumorous regions in images, as they are independent of human judgment and are reproducible [7].

Brain segmentation, particularly in neuroimaging, has gained significant attention and has become an active area of research [8]–[12]. Segmentation methods are generally categorized into generative models and discriminative models [6]. Generative models rely on domain-specific knowledge about brain tissue appearance and anatomy, while discriminative models depend on extracting numerous low-level features, such as local histograms or texture patterns [6].

The core principle of segmentation is to exploit intensity and texture variations among pixels [13]. This principle underlies various semi-automatic and fully automatic segmentation techniques, including thresholding, region-growing, and clustering. Hybrid methods that combine these techniques are also common. Examples of semi-automatic methods include support vector machines and neural networks, while automated techniques represent the current gold standard [6], [11], [14].

Automated brain tumor segmentation faces challenges due to the reliance on luminosity differences between tumorous and normal tissue pixels. These variations introduce complexities for automated systems and are further exacerbated by inter- and intra-expert variability in manual segmentation [6]. Additionally, differences in imaging modalities and patient data contribute to these challenges.

The remainder of this paper is organized as follows: Section II reviews related work. Section III outlines the methodology. Results and discussion are presented in Section IV, followed by the conclusion and future work in Section V.

## II. RELATED WORK

A substantial body of literature explores model checking as a technique for various applications. In [15], the authors provide a comprehensive survey and historical account of algorithmic requirements for directed model checking. Their work covers a wide range of topics, including bug-hunting techniques to mitigate the state explosion problem, prioritization of successor selection, and the adaptation of algorithms to time-based automata and probabilistic domains. Similarly, [16] focuses on algorithmic verification in probabilistic model checking, discussing various probabilistic models and their associated algorithms, with an emphasis on their reliability and dependability.

The extensive review in [17] highlights model-checking technology as a powerful approach for the automatic verification of hardware systems. The authors identify other application domains, translating verification problems into appropriate model-checking questions. Additionally, [17] introduces a taxonomy of models, properties, and model-checking approaches. In [18], the formal verification of statecharts using model checking is reviewed. A key observation in this work is the reliance of many statechart approaches on translating hierarchical structures into flat representations of the input language, which poses scalability challenges due to exponential growth in the state space.

Model checking for programmable logic controllers (PLCs) is comprehensively reviewed in [19], which highlights the importance of real-time functionality in production settings. Verification of PLC software using model checking is deemed critical in these scenarios. A theoretical overview by [20] examines the practical applications of model-checking schemes for verifying multi-threaded software systems, with a focus on the automata-theoretic method of verification and its associated challenges. Additionally, [21] surveys automata-theoretic approaches, focusing on the verification of probabilistic finite-state systems concerning linear-time properties.

Symbolic model checking is the subject of a detailed review in [22], which discusses critical system design and model-checking techniques within the domain of information sciences. The review by [23] explores the verification of web services using model checking, providing insights into data flow, requirements, and quality of service. This systematic review encompasses fifteen years of literature on web service verification. Finally, [24] reflects on the authors' experiences applying model checking to verify the arbitration logic of a vehicle control system, identifying strengths and limitations of different model-checking techniques and tools.

### A. Spatio-temporal Model Checking for Medical Imaging

The application of model checking frameworks in a spatial setting for medical imaging is a relatively new area of research, as evidenced by the limited number of publications in the field. However, the existing works, though few, are comprehensive. For tumor detection, machine learning has been utilized to extract image features and validate them using spatio-temporal models [25], [26].

In [27], spatio-temporal meta-model checking was employed for analyzing biological processes, with a strong emphasis on multi-scale aspects.

When it comes to fully automated approaches, machine learning-based techniques, particularly deep learning, have had a profound impact. Deep learning is highly effective in modeling the non-linearities inherent in data to derive meaningful insights [28].

For in vivo images, manual segmentation remains a standard practice [13]. However, it has several disadvantages. Manual segmentation demands significant focus and time from clinicians, making the process labor-intensive and expensive. Additionally, it requires a high level of expertise and is prone to subjectivity, resulting in variability among clinicians and introducing errors into the pipeline [13].

While machine learning and deep learning approaches have demonstrated remarkable success in detection, classification, and pattern recognition tasks, their effectiveness heavily depends on the quality and reliability of datasets. These frameworks often require large datasets capable of capturing sufficient variability. The strength of deep learning lies in its ability to extract features across multiple layers of raw data and make meaningful inferences. However, employing these techniques in a supervised framework necessitates ground truth labeling of the data, which introduces challenges. Ground truth annotations rely on clinicians' interpretations, making them subjective and variable among annotators. A study [13] reported intra-expert and inter-expert variability of up to 35% and 30%, respectively, in manually segmenting tumor regions in brain MRI images [29]. Consequently, there is a need for interactive model-checking methods to improve the reproducibility and efficiency of ground truth generation.

Spatio-temporal model checking is still a nascent area of research with limited literature available. Nonetheless, certain studies are noteworthy. For example, [30] implemented a spatial extension of signal temporal logic, which is linear and time-bounded, to provide a framework for stochastic population models. This framework is particularly suitable for scenarios where agents can move within a discrete spatial representation. Signal temporal logic employs a graph-based approach with finite cost-based weighted connections and a single spatial operator [13], similar to the operator introduced in [31]. Subsequent extensions, such as the bounded surrounded operator, enhanced the framework by adding new requirements, building on previous work [32]. In [33], this operator was further characterized as a derivation from a basic operator.

In [34], the author proposed an alternative approach by integrating spatial, statistical, and algorithmic frameworks instead of relying solely on spatial logics. This approach introduced distance and texture operators alongside spatial operators. Distance operators are often defined using real numbers and specific semantics. One example is the 'doughnut operator,' which meets distance constraints between two limits. [34] also explored quasi-discrete closure models defined by binary relations on a set of points, proposing the inclusion of distance operators to address the lack of distance information. In medical imaging, distances are typically based on Euclidean metrics or symmetric graphs. The concept of shortest path distances in weighted graphs is intuitive and useful. Other approaches involve sampling spatial grids, such as two-dimensional spaces with four or eight neighboring connections. However, variations in voxel dimensions pose inherent challenges in medical imaging.

Texture analysis operators are designed to detect and analyze patterns in medical images, many of which are imperceptible to the human visual system due to their subtle characteristics in brightness, color, or size [35]. Texture-based schemes are widely used in medical imaging [36], proving instrumental in diagnosing prostate cancer [37] and analyzing tumor heterogeneities [38].

B. Techniques used with model checking for analyzing the medical image

Many techniques are used with Model Checking in the analysis of medical images, the most important of which is Texture analysis, which has also found its application in dynamic contrast-enhanced MRI of breast imaging for the detection of and segmentation of malignant lesions [39]. In addition to that, there is extensive work on computer-aided diagnosis and segmentation using texture-based analysis [40], including diagnosis of pulmonary nodules [41]. Classification and segmentation using texture-primitive features in medical imaging have historically been an area of in-depth exploration. Extensive works of [42], [43], [44] have demonstrated the use of texture analysis for classification.

Texture analysis requires that the image textures be characterized by some quantitative measure. For that reason, certain descriptors are estimated to quantify these textural features [34]. A typical classification of these types of features includes syntactic, spectral and statistical [36].

In [34], the authors focused on first order statistical features primarily which entails extracting certain statistical descriptors from the distributions of features of each voxel. These first-order statistical features mainly consisted of statistics based on probability density functions of the intensity of voxels in the image. These could then be estimated as histograms by binning voxels values of the same intensities. Statistical features on the first order include the mean, variance, skewness, kurtosis, and entropy [45]. The advantage of using these statistical features, specifically in medical imaging is that they are invariant to transformations of the image. By construction, these first-order statistical operators are invariant to affine transformations that consist of rotation and scaling. This kind of transformation is critical in medical applications due to the variance involved in different image acquisition conditions. However, while this variance of first-order statistical descriptors is plausible, they also have a critical limitation which is their lack of spatial coherence. The features assume a degree of independence by ignoring the relative spatial placement of the voxels in the image. In the experimentation done in [34], the authors defined a logical operator to compare the areas of the image with a degree of statistical similarity to a pre-determined area. The idea is to search for sub-areas in the image with an empirical distribution similar to the predetermined area. For that purpose, some surrounding areas are also considered and a threshold is applied so as to obtain a Boolean value that confirms the voxel's statistical similarity to the sub-region. A measure for comparison for a statistical distribution used is cross-correlation [34]. As a result, the authors generalize the classical texture analysis making use of some spatial information owing to the examination of neighborhood distribution when investigating a particular voxel. The framework is implemented by the authors on an MR image slice of a brain affected by glioblastoma tumor [34]. In [13], the authors proposed an approach for segmentation based on a spatial logic method. The goal was to identify a region of interest in MR images for the analysis of glioblastoma as well as other tumors. The authors employed a texture-based method along with local histograms to create a hybrid approach for segmentation, maintaining relative spatial information. This is primarily inspired by a topological approach interwoven with spatial logic.

In this domain of research, works of [32] and [46] have been significant as they developed the theory to enhance the use of arbitrary graphs as models of space, particularly by utilizing a generalized form of topological spaces model called 'closure spaces' [47]. As a consequence of this, the spatial logic for closure spaces was formally defined along with a model-checking algorithm associated with it.

Another important application of model checking in medical imaging is the contouring of nevus images. Nevus is a visible, circumscribed lesion of the skin. Typically, it is small and benign. However, these are very difficult to distinguish from the malignant counterpart [48] which is medically known as a melanocytic nevus. Melanoma is an advanced form of this and is a serious form of skin cancer. As with most types of cancers, this too requires early detection and diagnosis for effective disease management and treatment. In the case that melanoma is not recognized at an early stage, it can become lethal and life-threatening. Reportedly, there are over 20,000 deaths due to melanoma in Europe each year [49]. Another critical factor that is the number of biopsies needed. This number can be cut down if automated systems are present to detect and diagnose melanoma. In the work of [48], the authors focused specifically on the contouring of the 2D images of nevi. This is generally considered to be a challenging problem since there is a lot of homogeneity in the texture, size, color, and shape of the nevi. Adding to the variance of the types of nevus, there are often extraneous elements that include patches, rulers, or hair. In [48], the authors take up this challenge by leveraging texture similarity operators along with spatial logic operators. Altogether, the feasibility of such a technique on the dermoscopic images from a public database is investigated in [48]. Usually, in the diagnosis of melanoma, segmentation of nevi is considered a part of the overall bigger problem. In the literature, [50] [51] [52] [13], there is evidence that techniques that include automatic contouring with spatial model checking for brain tumors have in fact shown quality comparable to the current state of the art. However, in the case of nevi segmentation and model checking, the additional challenges make it relatively complex, largely due to the optical effects that are associated with it. For instance, the type of lesions and contrast and the variance within the types of lesions.

Certain spatial model checkers essentially utilize high-level specifications written in a logic language in order to describe certain spatial properties so as to identify spatial patterns of interest in an automatic and efficient way. These spatial patterns are key in identification of spatial structures of [48]. The images are conveniently described by their voxels. In [48], the authors investigate the feasibility of a technique that is based on spatial logic for closure spaces for the analysis of nevi images from a public data set. The authors show that despite the in homogeneity in the nature of nevi, ranging from variance in shape, color, texture, and size to inclusion of extraneous elements like rulers, patches, and hair, the authors were able to demonstrate the analytic ability in a semi-automatic way. This is credited to the intrinsic rigor of a logic-based approach, utilizing the efficient implementation of spatial model-checking algorithms. The comparison is then drawn against the ground truth that is provided along with the public dataset. The dataset used is the one provided by the Skin Lesion Analysis toward Melanoma Detection challenge 2016 [53]. The composition of the data set includes high-resolution annotated dermoscopic images. The ground truth segmentation data is done manually by the experts and are separately available as ground truth mask images.

[48] also reports variance not only in the different nevi but also within one nevus. The change of texture and color within a nevus poses a great segmentation challenge. This is even more complicated with the presence of any sebaceous follicle. Since the inter and intra variations in the nexus pose a great challenge in segmentation, the approach is then tailored to focus primarily on the differentiation between any type of nevus tissue and skin tissue. This is done with the help of texture analysis as well as spatial operators so as to isolate the nevus tissue from the skin tissue. To that end, the authors employ a statistical texture analysis operator to approximate a nevus. This is done by using a statistical heuristics approach to distinguish between background skin and likely parts of nevus. There is an underlying assumption that the likely part of nevus is centered in the image so that the area at the periphery of the image belongs to the background healthy skin cells [48]. This allows for the algorithm to take a sample of the background. Here, only the intensities are considered while hue or saturation is ignored. Using these intensities only, a histogram of the distribution of intensities is computed with a suitable number of bins. While there is a global histogram, one that takes into account all the pixels in the image, a smaller, more localized histogram is also constructed for each pixel where intensities of surrounding pixels in a specific radius are considered. Comparison of these local histograms with the global histogram in the form of a Pearson correlation coefficient allows for a comparison of the nature of the localized areas. This texture operator serves as a good first approximation method to identify the area covered by the nevus. However, this is still an approximation and requires adjustments to be made, which are subsequently performed using derived operators with metrics like relative distance and similarity indexes. The spatial model checking techniques can be used with the spatial model checking tool VoxlogicA developed in [51] to efficiently segment nevi. The segmentation based on dermoscopic images is important in automatic routines for the diagnosis of malign skin tumors including

but not limited to melanoma. Spatial model checkers make use of high-level logic languages to identify certain spatial properties. In conclusion, [48] presented a novel segmentation method to combine spatial operators, that were inspired by closure spaces, and domain-based operators such as the texture similarity operator. As a result, achieving a dice score of 0.9 validates the segmentation quality. One of the outcomes of employing such a technique is that it is explainable in its methods and it is mostly high-level [48]. There is room to advance this work by increasing the number of classes of images for which the segmentation has shown remarkable accuracy. It is pertinent to mention the scale of homogeneity between the nevi and within the nevi as well. Such variations reflect artifacts in the image and pose a tremendous challenge for accurate segmentation.

C. The Spatial Logic Framework: VoxlogicA and SQL

One of the most important frameworks for image analysis using the model Checking is VoxLogic, which is a framework for image processing that incorporates user-oriented expression languages into the logic ImgQL to edit images[51], this tool takes advantage of the library of computational imaging algorithms alongside distinct combinations of the declarative specification to deliver optimized execution inherent to the spatial logic model checking. As a consequence, the methodology developed is considered to be rapid. Testing this methodology on existing brain tumor segmentation benchmark images shows that the accuracy can reach the state of the art. The additional advantage is the explainability and replicability of the approach [51].

The fundamental idea of Spatio-temporal model checking is to use the specifications in a relevant logical language in order to describe the spatial characteristics so that patterns and structures of key importance can automatically be identified. In [51], the main focus is on medical imaging for radiotherapy, particularly, brain tumor segmentation. A challenge in this domain is that the tumorous regions or lesions are only defined distinctively from the normal tissue, owing to any changes in the intensities of the pixels in the gray-scale images. This relativity of pixel intensity as a marker of tumor presence makes it a complex challenge to isolate the lesions from normal tissue pixels. In addition to that, there is a considerable variation in the ground truth images of these segmented images. This is due to the variance in the manual segmentation by the experts. When there are intensity gradients between adjacent tissue structures, the experts have shown significant subjectivity in the assessment of ground truth for segmenting tumors [51]. This adds to the already challenging constraint of isolating tumors from normal tissues in gray-scale images based on intensities as a metric. Furthermore, the inconsistency in the data complicates things further as different MRI scanners also show considerable variation in the image quality. However, for demonstration of the approach, the authors use the publicly available BraTS 2017 dataset [6] where ground truth data is available for any objective deductions; reporting accuracies competitive with the state-of-the-art techniques for glioblastoma segmentation.

The authors build on the image query language(SQL) that was proposed in [13]. This was in turn based on the spatial logic for closure spaces [46]. It is from [46] that the authors in [51] derive their kernel for their framework. The work in [51] is closely related to the spatial logic for closure spaces presented in [13], particularly with regard to the distance-based operator formed therein. For the digital image analysis, a statistical similarity operator is used that quantifies the similarity of an area around a point with that of a given region. This is achieved by the computation of respective histograms and then finding cross-correlation between them. This operator allows checking to what extent the area around a point of interest is statistically similar to a given region.

Another operator introduced is the percentile operator which takes a numeric-value-based image and its binary mask in order to return an image that shows at how each point is associated to the percentile rank of its intensity with respect to the population voxels. This allows for the same segmentation specification on images that have different intensity distributions. This also allows for avoiding the use of absolute values in constraints on the intensity of points. In terms of its functionality, the VoxLogicA tool specializes for spatial analysis for multidimensional images. The pipeline for VoxLogicA is fairly simple as it interprets a particular specification that is written in the image query language and produces a set of multidimensional images representing the valuation of user-specified expressions. However, when it comes to medical images, the images are boolean-valued for logical operators, thus, the regions of interest may be overlaid on top of the original images for better viewing. On the other hand, the non-logical operators will yield number-valued images. The tool was used in [51] for evaluation in two ways. First, the evaluation was done for VoxLogicA for the segmentation of Glioblastoma in medical images obtained from MRI scans where the clinical target volume of the whole tumor is considered. This is different from the gross tumor volume which corresponds to what is actually visible on an image. Secondly, the evaluation was done on the BraTS 2017 dataset which serves as a quality measure of the proposed technique. The Dice similarity metric serves as a means to quantify the segmentation accuracy by computing a numeric value for the overlap between the manually segmented ground truth and the ones achieved by the VoxLogicA tool. The Dice result of 0.9 reflects the tool is sufficiently accurate [51].

The evaluation of VoxLogicA involved sets of 3D images of size $240 \times 240 \times 155$ which is about 9 million voxels. Using a variant of [13] a comparison is made to assess the performance of VoxLogicA and another model, topochecker. The specifications consist of two human-authored text files of about 30 lines each, identifying the oedema. The machine used for testing is a desktop computer equipped with a 7th-generation Intel Core I7 processor and 16GB of RAM. In the 2D case (image size: $512 \times 512$), topochecker takes 52 seconds to complete the analysis, whereas VoxLogicA takes

750 milliseconds. In the 3D case (image size: 512 × 512 × 24), topochecker takes about 30 minutes, whereas VoxLogicA takes 15 seconds. This huge improvement is due to the combination of a specialized imaging library, new algorithms such as statistical similarity of regions, parallel execution, and other optimizations.

In [52], the focus has largely been on identifying tissues in the healthy brain. For example, the idea is to determine white matter or gray matter in a healthy brain.
This is different from segmenting and identifying tumorous regions in the brain as done in other studies. [52] reiterates that model checking in medical imaging is mostly concerned with the creation of visual representations of parts of the human body for clinical analysis as well as preparedness for medical intervention. A key step in radiotherapy planning includes contouring tissues and organs accurately. This, in conjunction with automatic contouring simplifies diagnostic procedures and eventually contributes to the reduction of time and cost as compared to manual contouring. The automated software that is used for contouring is typically highly specific in its application, for instance, having the ability to only contour the human brain [52]. Such software offers little flexibility and transparency for the users and often does not deliver satisfactory accuracy [52]. Deep learning-based approaches have recently become popular for medical image analysis. These are computationally efficient and mostly deliver decent results. However, they do require sufficiently large data sets that are accurately labeled by experts. Such large datasets are often not available, thus, posing as a limiting factor to these deep learning-based techniques. In addition to that, there is inter and intra variability between the labelers who annotate the data for ground truth. In [52], the focus is on healthy brain images to identify the types of tissues. The framework for the spatial logic utilized by [52] involves modeling a digital image as an adjacency space. This effectively means that only the pixels that share an edge are counted as adjacent. Different types of adjacency exist due to the nature of the rules of adjacency. For example, an orthodiagonal adjacency scheme considers a pixel adjacent even if a corner is shared. Since each pixel is associated with color intensities, this can be modeled for similar pixels by using another metric called attributes. These attributes can be passed onto an attribute function and be used for Boolean expressions with threshold parameters. This aids in defining a closure space. In effect, the adjacency space is a subclass of closure spaces. The closure spaces can be strengthened using distance metric, thus, leading to distance closure spaces.

In the work of [52], certain new derived operators have been introduced for the spatial model checking framework. Touch is a derived operator that guarantees that a starting point and an ending point meet particular criteria along with all intermediate points in between them. With particular criteria set, a grow operator is also defined. A filter operator functions as a filter, taking a radius of a certain distance into account, ultimately resulting in a smoothing operation. Furthermore, a statistical similarity operator is also defined. It searches for tissues that have similar textural characteristics by comparing the similarity of the histograms of relevant regions. This statistical similarity-based operator is a cross-correlation-based operator. This operator is also invariant to rotations which adds to this importance for use in spatial logic in medical applications. A demonstration on the benchmark checkerboard pattern shows good performance. The authors in [52] illustrate the brain segmentation on two simulated images of the brain. Simulated images provide the benefit of conclusive ground truths [52]. In this particular application, this is even more critical as the regions of interest in question are perfectly normal brain tissues; the only difference is in the nature of the brain tissue, such as the gray matter and the white matter. For quantitatively testing a method, using simulated images is, therefore, very effective.

In the above-discussed literature, the works are aimed at reviewing model checking as a verification tool set in a multitude of domains. However, at the time of writing this work, there is no review or survey that explores the specific application of medical imaging with model checking as a means of verification, particularly with spatial logic design. This paper is a first of its kind reviewing different model-checking approaches taken to understand segmentation and identification of regions of interest in medical images.

### III. METHODOLOGY

This work reviews recent research papers focusing on the application of model checking in medical imaging. A comprehensive analysis was conducted by compiling a curated list of relevant studies. To achieve this, multiple web-based scientific databases were accessed using keywords such as "model checking" and "medical imaging." The review highlights key methodologies employed in these studies, particularly those utilizing model checking to validate proposed techniques. Special emphasis is placed on tumor segmentation within medical images, a prominent area of research demonstrating significant potential and promise. This study aims to critically evaluate and synthesize findings from the most recent works employing model checking for segmentation tasks.

### IV. RESULTS AND DISCUSSION

In this section, we review the findings of some of the works on model checking in medical imaging and build our discussion on top of these works. An analysis of these works in the form of a compilation is also shown in this section so as to discuss these works in detail. While there are conclusive advantages to using spatial model-checking methods for medical imaging, it is imperative to understand the associated challenges.

Logic-based techniques in model checking have a strong reliance on basic set of logical operators. These logical operators allow for the definition of more expressive derived operators that resonate with the domain-specific reasoning of the user. The idea is to exploit the compositionality of the basic operators so as to build more complex operators. A benefit for using very basic building blocks is the underlying

| Sr | Studies | Application | Innovation/Paper-Type | Result | Year |
| --- | --- | --- | --- | --- | --- |
| 1 | Karmakar et al.[22] | System Design | Review | -Algorithmic verification of probabilistic finite-state systems with respect to linear-time properties. | 2022 |
| 2 | Belmont et al.[48] | Nevus Segmentation | Spatial MC | -Determination of the contour of the two-dimensional images of nevi. | 2021 |
| 3 | Gopal et al.[23] | Web Service | Review | -A systematic review of the current research work on web service validation-based model validation that emerged during the period 2002-2017. | 2021 |
| 4 | Jonas et al.[24] | Vehicle Control | Review | -Report on model check application trials to check the arbitration logic of the vehicle control system. Identify the pros and cons of different model-checking techniques and tools | 2020 |
| 5 | Vincenzo et al.[51] | Glioblastoma | VoxLogicA | -Automatic segmentation of glioblastoma in MR flair for radiotherapy | 2019 |
| 6 | Massink et al. [52] | Glioblastoma | Spatial MC | Providing an open platform introducing declarative medical image analysis. | 2019 |
| 7 | Buonamici et al. [13] | Glioblastoma | ImgQL | Introducing the logical language ImgQL ("Image Query Language"). ImgQL extends SLCS with Boolean operators that describe distance and area similarity. | 2018 |
| 8 | Belmont et al.[50] | Glioblastoma | Automatic Segmentation | -Automatic segmentation of glioblastoma | 2017 |
| 9 | Nenzi et al. [30] | Reaction Diffusion | Spatio-temporal Logic | -Introducing the Signal SpatioTemporal Logic (SSTL), a modality that may be utilised to characterise the spatiotemporal characteristics of linear time and discrete space models. | 2016 |
| 10 | Katoen et al. [16] | Probabilistic MC | Review | Algorithmic verification of probabilistic models, in particular probabilistic model checking. | 2016 |
| 11 | Ovidiu et al.[27] | Biological Systems | Meta MC | A New Approach to Multiscale Spatio-Temporal Meta Model Checking for Multilevel Computational Models of Biological Systems | 2016 |
| 12 | Ovatman et al[19] | PLC | Review | Model checking practices on verification of PLC software," Software and Systems Modeling | 2016 |
| 13 | Ciancia et al.[32] | Spatial Operator | Topological Generalization | Defining and confirming space attributes | 2016 |
| 14 | Belmonte et al.[34] | Medical Imaging | Multiple Operators | Presenting a preliminary experiment focusing on spatial model verification applications to MI. | 2016 |
| 15 | Edelkamp et al.[15] | State-explosion | Review | Using Histological Image Processing Techniques, Chronic Tumor Hypoxia is Quantitatively Characterized | 2008 |
| 16 | Gerard et al.[20] | Software systems | Review | Explaining the theoretical underpinnings and practical applications of logic model-checking methods for the verification of multi-threaded software. | 2005 |
| 17 | Bhaduri [18] | State-chart models | Review | Recommending a variety of approaches to deal with the issue of state space explosion as well as some future research topics. | 2004 |
| 18 | Vardi et al.[21] | Probabilistic MC | Review | Being able to reduce probabilistic model checking for ergodic analysis of Markov chains. | 1999 |
| 20 | Reif and al.[31] | Packet Routing | Spatio-temporal Operators | spatial and temporal modalities in a multiprocess network logic | 1985 |

efficiency for the verification and correctness of the algorithms. It is also expected that the approach has some degree of flexibility and generalization so that it does not unnecessarily be coupled with a particular application. Furthermore, a lot of focus in medical imaging and diagnosis has been on deep learning-based techniques where hierarchical features are extracted and a model is trained to fit under plenty of hyperparameters to give desired results. However, this requires large datasets and with significant variability in inter and intra experts on contours of segmentation, not only is deep learning implementation time-consuming for manual segmentation help improve as well. To that end, interactive approaches based on spatial model checking may be of help in terms of improving the generation of manual ground truth annotations in a more efficient, transparent, and reproducible way. The explaining ability aspect is also an under-explored problem when it comes to deep learning methods as deep learning approaches suffer from a lack of reasonable explanations on human insight as to why a certain area is classified as a tumor by the model, particularly when the models do not provide correct results at all times. The results in [52] is obtained by utilizing the VoxLogicA-based model checking approach with the integration of the above-discussed derivative operators.

The quantitative metrics used for evaluation included the dice coefficient, specificity, and sensitivity. Sensitivity quantifies the fraction of pixels correctly identified as the ground truth whereas specificity quantifies the fraction of pixels identified as true negatives. The dice index encapsulates the similarity. The authors were able to achieve dice, sensitivity, and specificity scores as high as 0.90, 0.91, and 0.98, respectively for grey matter in the brain MRI image and 0.89, 0.85, and 1.0, respectively for white matter. In [48] work on nevus segmentation, the authors achieved dice, sensitivity, and specificity scores of 0.81, 0.81, and 0.96, respectively. While some variation on scores is evident owing to the difference in applications, the objectivity in the strong performance of model checking as a tool for segmentation, in general, is appreciable nonetheless. [52] emphasized the importance of enhancing capabilities and ensuring transparency in annotation procedures conducted by experts during segmentation. Given their extensive experience in marking ground truths, it is vital to encapsulate this expertise in the form of high-level operations with formal specifications. These specifications should be exchangeable, publishable, and open to discussion within the domain expert community, promoting collaboration and driving advancements in the field.

Furthermore, ensuring device independence is essential for the broader adoption of academically significant research in clinical practice. Several factors hinder this transition, including the lack of integration of academic research outcomes with existing hardware and the reliance on hard-coded execution environments. To facilitate successful technology transfer, it is critical to address the fragmentation of academic research and establish open standards and protocols. These measures can enable intermediaries, such as imaging device vendors, to transform innovative ideas into practical clinical applications. Additionally, challenges related to data privacy and procedural confidentiality underscore the need for open standards that allow users to exchange analysis procedures and build a foundation of shared knowledge. We summarize the contributions of the relevant literature on model checking in Table 1.

The table results show that most of the publications we studied focused primarily on segmentation and analysis of medical images. This is to confirm the effectiveness of the model checking in the analysis of medical images, as the accuracy of most of these works exceeded 90%, which is a percentage that shows the importance of model checking in the analysis of medical images, in order to learn more about the areas of use of model checking (computer science, engineering, Medicine Public Health, medicine ), figure 1 shows the results of the top 5 areas where the model checking was applied by relying on Spring Nature statistics.

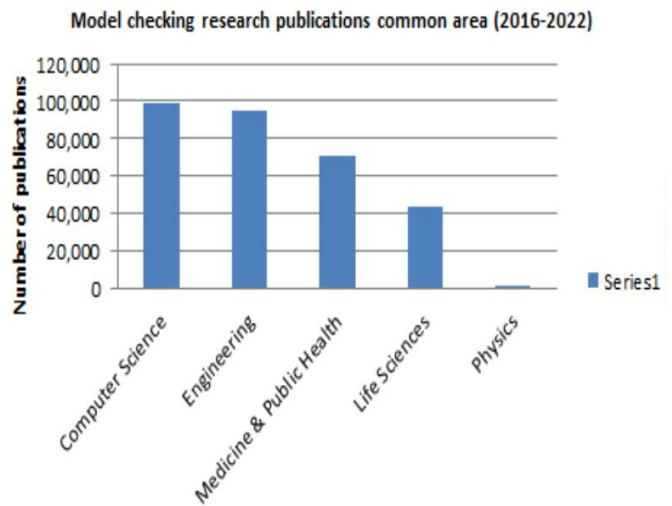

Figure 2: Model checking research publications common area

In order to be more precise, we calculated the percentage for each field in terms of the number of Model Checking publications by relying on Spring Nature statistics.

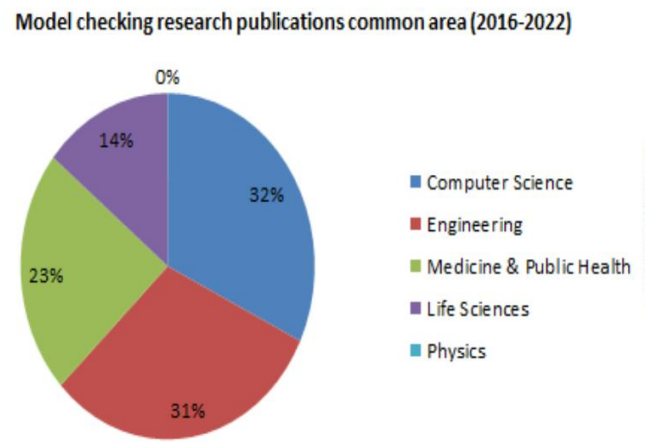

Figure 3: Model checking research publications common area

It is clear from the above diagram that the strength of model checking is the ability to apply it in different areas, unlike some other methods, and it is clear that most applications of model checking are in computer engineering sciences due to the need for computer engineering to verify the correctness of the characteristics in order to avoid errors in the development of programs.

we used the systematic review to draw the historical publication chart of model-checking research publications per year

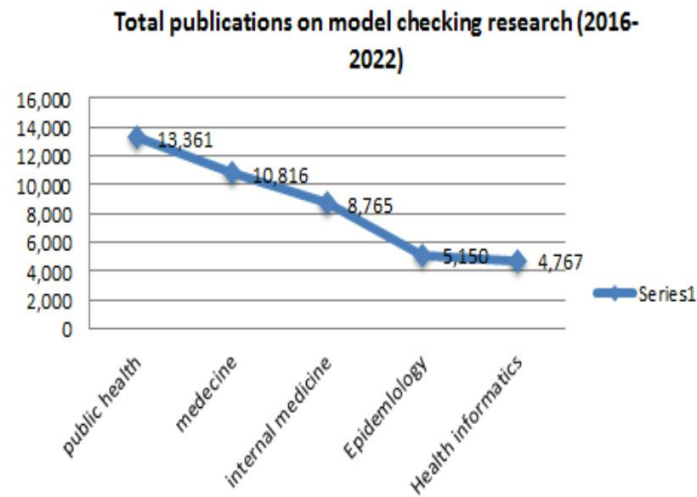

Figure 4: Total publications on model checking research (2016–2022)

We note that the importance of model checking in the health field is being studied due to the ability of the model checking to identify and avoid errors, in addition to its low cost compared to deep learning techniques

## V. Conclusion and future work

Model checking in medical imaging remains an active area of research. The use of spatial model checking for verifying and validating spatial logic designs plays a critical role in identifying and segmenting various types of tissues in the body. However, significant challenges persist, particularly in the technical development of frameworks and the explainability and reproducibility of methodologies within the model-checking domain. The existing literature demonstrates that model checking is an effective tool not only for detecting tumorous tissues but also for distinguishing between different healthy tissues with notable accuracy. Despite these achievements, there remains considerable room for improvement, as much of the existing research focuses on specific applications. Moving forward, we aim to advance model-checking techniques for medical image analysis, with a particular focus on applying these methods to 3D imaging.

## Declarations

- Funding: This research received no external funding.
- Conflict of interest: The authors declare no conflict of interest.
- Informed Consent Statement: Not applicable.
- Data Availability Statement: Not applicable.
- Research Involving Human and /or Animals :Not applicable

## References

[1] B. Halalli and A. Makandar, "Computer aided diagnosis - medical image analysis techniques," in Breast Imaging, C. M. Kuzmiak, Ed. Rijeka: IntechOpen, 2017, ch. 5. [Online]. Available: https://doi.org/10.5772/intechopen.69792

[2] N. Gordillo, E. Montseny, and P. Sobrevilla, "State of the art survey on mri brain tumor segmentation," Magnetic Resonance Imaging, vol. 31, no. 8, pp. 1426–1438, 2013.

[3] K. K. Brock, Image Processing in Radiation Therapy. CRC Press, 2013.

[4] G. Chetelat and J. C. Baron, "Early diagnosis of alzheimer's disease: contribution of structural neuroimaging," NeuroImage, vol. 18, no. 2, pp. 525–541, 2003.

[5] A. Toosy, D. Werring, R. Orrell, R. Howard, M. King, G. Barker, D. Miller, and A. Thompson, "Diffusion tensor imaging detects corticospinal tract involvement at multiple levels in amyotrophic lateral sclerosis," Journal of neurology, neurosurgery, and psychiatry, vol. 74, no. 9, p. 1250—1257, September 2003.

[6] B. Menze, "The multimodal brain tumor image segmentation benchmark (brats)," IEEE Transactions on Medical Imaging, vol. 34, no. 10, pp. 1993–2024, 2015.

[7] C. Dupont, N. Betrouni, N. Reyns, and M. Vermandel, "On image segmentation methods applied to glioblastoma: State of art and new trends," IRBM, vol. 37, no. 3, pp. 131–143, 2016.

[8] A. E. Lefohn, J. E. Cates, and R. T. Whitaker, "Interactive, gpu-based level sets for 3d segmentation," in International Conference on Medical Image Computing and Computer-Assisted Intervention. Springer, 2003, pp. 564–572.

[9] S. Ho, E. Bullitt, and G. Gerig, "Level-set evolution with region competition: automatic 3-d segmentation of brain tumors," in 2002 International Conference on Pattern Recognition, vol. 1. IEEE, 2002, pp. 532–535.

[10] M. Wels, G. Carneiro, A. Aplas, M. Huber, J. Hornegger, and D. Comaniciu, "A discriminative model-constrained graph cuts approach to fully automated pediatric brain tumor segmentation in 3-d mri," in International Conference on Medical Image Computing and Computer-Assisted Intervention. Springer, 2008, pp. 67–75.

[11] S. Bauer, J. Tessier, O. Krieter, L.-P. Nolte, and M. Reyes, "Integrated spatio-temporal segmentation of longitudinal brain tumor imaging studies," in International MICCAI Workshop on Medical Computer Vision. Springer, 2013, pp. 74–83.

[12] A. Hamamci, N. Kucuk, K. Karaman, K. Engin, and G. Unal, "Tumor-cut: segmentation of brain tumors on contrast enhanced mr images for radiosurgery applications," IEEE transactions on medical imaging, vol. 31, no. 3, pp. 790–804, 2011.

[13] F. B. Buonamici, G. Belmonte, V. Ciancia, D. Latella, and M. Massink, "Spatial logics and model checking for medical imaging," 2018.

[14] G. Mohan and M. M. Subashini, "Mri based medical image analysis: Survey on brain tumor grade classification," Biomedical Signal Processing and Control, vol. 39, pp. 139–161, 2018.

[15] S. Edelkamp, V. Schuppan, D. Bošnački, A. Wijs, A. Fehnker, and H. Aljazzar, "Survey on directed model checking," in Model Checking and Artificial Intelligence, D. A. Peled and M. J. Wooldridge, Eds. Berlin, Heidelberg: Springer Berlin Heidelberg, 2009, pp. 65–89.

[16] J.-P. Katoen, "The probabilistic model checking landscape," in Proceedings of the 31st Annual ACM/IEEE Symposium on Logic in Computer Science, ser. LICS '16. New York, NY, USA: Association for Computing Machinery, 2016, p. 31–45.

[17] M. Müller-Olm, D. Schmidt, and B. Steffen, "Model-checking," in Static Analysis, A. Cortesi and G. Filé, Eds. Berlin, Heidelberg: Springer Berlin Heidelberg, 1999, pp. 330–354.

[18] P. Bhaduri and S. Ramesh, "Model checking of statechart models: Survey and research directions," 2004.

[19] T. Ovatman, A. Aral, D. Polat, and A. O. Ünver, "An overview of model checking practices on verification of PLC software," Software and Systems Modeling, vol. 15, no. 4, pp. 937–960, oct 2016.

[20] G. J. Holzmann, "Software model checking with spin," ser. Advances in Computers. Elsevier, 2005, vol. 65, pp. 77–108.

[21] M. Y. Vardi, "Probabilistic linear-time model checking: An overview of the automata-theoretic approach," in Formal Methods for Real-Time and Probabilistic Systems, J.-P. Katoen, Ed. Berlin, Heidelberg: Springer Berlin Heidelberg, 1999, pp. 265–276.

[22] R. Karmakar, "Symbolic model checking: A comprehensive review for critical system design," in Advances in Data and Information Sciences, S. Tiwari, M. C. Trivedi, M. L. Kolhe, K. Mishra, and B. K. Singh, Eds., 2022, pp. 693–703.

[23] G. N. Rai and G. R. Gangadharan, "Model checking based web service verification: A systematic literature review," IEEE Transactions on Services Computing, vol. 14, no. 3, pp. 747–764, 2021.

[24] J. Fritzsch, T. Schmid, and S. Wagner, "Experiences from large-scale model checking: Verification of a vehicle control system," CoRR, vol. abs/2011.10351, 2020.

[25] A. Sundstrom, E. Grabocka, D. Bar-Sagi, and B. Mishra, "Histological Image Processing Features Induce a Quantitative Characterization of Chronic Tumor Hypoxia," PloS one, vol. 11, no. 4, 2016.

[26] R. Grosu, E. Bartocci, F. Corradini, E. Entcheva, S. A. Smolka, and A. Wasilewska, "Learning and detecting emergent behavior in networks of cardiac myocytes," in Hybrid Systems: Computation and Control, M. Egerstedt and B. Mishra, Eds. Berlin, Heidelberg: Springer Berlin Heidelberg, 2008, pp. 229–243.

[27] P. Ovidiu and G. David, "A Novel Method to Verify Multilevel Computational Models of Biological Systems Using Multiscale Spatio-Temporal Meta Model Checking," PloS one, vol. 11, no. 5, may 2016.

[28] Z. Akkus, A. Galimzianova, A. Hoogi, D. L. Rubin, and B. J. Erickson, "Deep Learning for Brain MRI Segmentation: State of the Art and Future Directions," Journal of digital imaging, vol. 30, no. 4, pp. 449–459, aug 2017. [Online]. Available: https://pubmed.ncbi.nlm.nih.gov/28577131/

[29] G. P. Mazzara, R. P. Velthuizen, J. L. Pearlman, H. M. Greenberg, and H. Wagner, "Brain tumor target volume determination for radiation treatment planning through automated MRI segmentation," International Journal of Radiation Oncology Biology Physics, vol. 59, no. 1, pp. 300–312, may 2004.

[30] L. Nenzi, L. Bortolussi, V. Ciancia, M. Loreti, and M. Massink, "Qualitative and Quantitative Monitoring of Spatio-Temporal Properties with SSTL," vol. 14, no. 4, pp. 1–38, 2017.

[31] J. Reif and A. Sistla, "A multiprocess network logic with temporal and spatial modalities," Journal of Computer and System Sciences, vol. 30, no. 1, pp. 41–53, 1985.

[32] V. Ciancia, D. Latella, M. Loreti, and M. Massink, "Specifying and verifying properties of space," in Theoretical Computer Science, J. Diaz, I. Lanese, and D. Sangiorgi, Eds. Berlin, Heidelberg: Springer Berlin Heidelberg, 2014, pp. 222–235.

[33] E. Bartocci, L. Bortolussi, M. Loreti, and L. Nenzi, "Monitoring mobile and spatially distributed cyber-physical systems," in Proceedings of the 15th ACM-IEEE International Conference on Formal Methods and Models for System Design. ACM, sep 2017.

[34] G. Belmonte, V. Ciancia, D. Latella, and M. Massink, "From collective adaptive systems to human centric computation and back: Spatial model checking for medical imaging," Electronic Proceedings in Theoretical Computer Science, vol. 217, pp. 81–92, jul 2016.

[35] G. Castellano, L. Bonilha, L. M. Li, and F. Cendes, "Texture analysis of medical images," Clinical radiology, vol. 59, no. 12, pp. 1061–1069, dec 2004.

[36] A. Kassner and R. E. Thornhill, "Texture analysis: a review of neurologic MR imaging applications," AJNR. American journal of neuroradiology, vol. 31, no. 5, pp. 809–816, may 2010.

[37] R. Lopes, A. Ayache, N. Makni, P. Puech, A. Villers, S. Mordon, and N. Betrouni, "Prostate cancer characterization on MR images using fractal features," Medical physics, vol. 38, no. 1, pp. 83–95, 2011. [Online]. Available: https://pubmed.ncbi.nlm.nih.gov/21361178/

[38] F. Davnall, C. S. Yip, G. Ljungqvist, M. Selmi, F. Ng, B. Sanghera, B. Ganeshan, K. A. Miles, G. J. Cook, and V. Goh, "Assessment of tumor heterogeneity: an emerging imaging tool for clinical practice?" Insights into imaging, vol. 3, no. 6, pp. 573–589, dec 2012.

[39] B. J. Woods, B. D. Clymer, T. Kurc, J. T. Heverhagen, R. Stevens, A. Orsdemir, O. Bulan, and M. V. Knopp, "Malignant-lesion segmentation using 4D co-occurrence texture analysis applied to dynamic contrast-enhanced magnetic resonance breast image data," Journal of magnetic resonance imaging : JMRI, vol. 25, no. 3, pp. 495–501, mar 2007.

[40] T. Heinonen, T. Arola, A. Kalliokoski, P. Dastidar, M. Rossi, S. Soimakallio, J. Hyttinen, and H. Eskola, "Computer aided diagnosis tool for the segmentation and texture analysis of medical images," in World Congress on Medical Physics and Biomedical Engineering, September 7 - 12, 2009, Munich, Germany, O. Dössel and W. C. Schlegel, Eds. Berlin, Heidelberg: Springer Berlin Heidelberg, 2009, pp. 274–276.

[41] F. Han, H. Wang, G. Zhang, H. Han, B. Song, L. Li, W. Moore, H. Lu, H. Zhao, and Z. Liang, "Texture feature analysis for computer-aided diagnosis on pulmonary nodules," Journal of digital imaging, vol. 28, no. 1, pp. 99–115, feb 2015.

[42] C.-C. Chen, J. DaPonte, and M. Fox, "Fractal feature analysis and classification in medical imaging," IEEE Transactions on Medical Imaging, vol. 8, no. 2, pp. 133–142, 1989.

[43] N. Sharma, A. Ray, S. Sharma, K. Shukla, S. Pradhan, and L. Aggarwal, "Segmentation and classification of medical images using texture-primitive features: Application of BAM-type artificial neural network," Journal of Medical Physics / Association of Medical Physicists of India, vol. 33, no. 3, p. 119, jul 2008.

[44] F. Tomita and S. Tsuji, Statistical Texture Analysis. Boston, MA: Springer US, 1990, pp. 13–36.

[45] G. N. Srinivasan and G. Shobha, "Statistical texture analysis," International Journal of Computer and Information Engineering, vol. 2, no. 12, pp. 4268 – 4273, 2008. [Online]. Available: https://publications.waset.org/vol/24

[46] V. Ciancia, D. Latella, M. Loreti, and M. Massink, "Spatial logic and spatial model checking for closure spaces," Lecture Notes in Computer Science (including subseries Lecture Notes in Artificial Intelligence and Lecture Notes in Bioinformatics), vol. 9700, pp. 156–201, 2016.

[47] A. Galton, "The mereotopology of discrete space," Lecture Notes in Computer Science (including subseries Lecture Notes in Artificial Intelligence and Lecture Notes in Bioinformatics), vol. 1661, pp. 251–266, 1999.

[48] G. Belmonte, G. Broccia, V. Ciancia, D. Latella, and M. Massink, "Feasibility of spatial model checking for nevus segmentation," in 2021 IEEE/ACM 9th International Conference on Formal Methods in Software Engineering (FormaliSE), 2021, pp. 1–12.

[49] A. M. Forsea, V. Del Marmol, E. De Vries, E. E. Bailey, and A. C. Geller, "Melanoma incidence and mortality in Europe: new estimates, persistent disparities," British Journal of Dermatology, vol. 167, no. 5, pp. 1124–1130, nov 2012. [Online]. Available: https://onlinelibrary.wiley.com/doi/full/10.1111/j.1365-2133.2012.11125.xhttps://onlinelibrary.wiley.com/doi/abs/10.1111/j.1365-2133.2012.11125.xhttps://onlinelibrary.wiley.com/doi/10.1111/j.1365-2133.2012.11125.x

[50] G. Belmonte, C. Vincenzo, L. Diego, and M. Mieke, "A topological method for automatic segmentation of glioblastoma in mr flair for radiotherapy," 2017.

[51] G. Belmonte, V. Ciancia, D. Latella, and M. Massink, "Voxlogica: a spatial model checker for declarative image analysis (extended version)," in TACAS, 2019.

[52] ——, "Innovating medical image analysis via spatial logics," in From Software Engineering to Formal Methods and Tools, and Back, 2019.

[53] N. C. F. Codella, Q.-B. Nguyen, S. Pankanti, D. Gutman, B. Helba, A. C. Halpern, and J. R. Smith, "Deep learning ensembles for melanoma recognition in dermoscopy images," ArXiv, vol. abs/1610.04662, 2017.